\pdfoutput=1
\documentclass[11pt]{article}

\usepackage{acl}

\usepackage{times}
\usepackage{latexsym}
\usepackage{multirow} 
\usepackage{array}
\usepackage[T1]{fontenc}
\usepackage[utf8]{inputenc}

\usepackage{microtype}

\usepackage{inconsolata}

\usepackage{graphicx}

\title{The Zeno's Paradox of `Low-Resource' Languages}

\author{\normalsize  Hellina Hailu Nigatu$^{1,}$ \thanks{ \ \ Work done while this author was at MBZUAI.} \hspace{1cm}  Atnafu Lambebo Tonja$^{2,3,}$ \\
\textbf{\normalsize Benjamin Rosman $^{3,4,}$ \thanks{\ \ These authors provided equal advice and supervision.} \hspace{1cm}  Thamar Solorio $^{2,5,}$ \footnotemark[2] \hspace{1cm}  Monojit Choudhury $^{2,}$\footnotemark[2]} \\
\footnotesize
Corresponding author: hellina\_nigatu@berkeley.edu \\\\
\footnotesize
$^1$ UC Berkeley, USA, $^{2}$ MBZUAI, UAE, $^{3}$ Lelapa AI, South Africa \\
 \footnotesize
$^4$ RAIL Lab - University of the Witwatersrand, South Africa, $^5$ University of Houston, Houston, USA
}
\begin{document}
\maketitle
\begin{abstract}
  The disparity in the languages commonly studied in Natural Language Processing (NLP) is typically reflected by referring to languages as low vs high-resourced. 
    However, there is limited consensus on what exactly qualifies as a `low-resource language.'
    To understand how NLP papers define and study `low resource' languages, we qualitatively analyzed 150 papers from the ACL Anthology and popular speech-processing conferences that mention the keyword `low-resource.'
    Based on our analysis, we show how several interacting axes contribute to `low-resourcedness' of a language and why that makes it difficult to track progress for each individual language. We hope our work (1) elicits explicit definitions of the terminology when it is used in papers and (2) provides grounding for the different axes to consider when connoting a language as low-resource.

\end{abstract}

\section{Introduction}
% \begin{quote}
%  [...] all ``low-resource'' languages should not be conflated together into one large group, but rather considered independently, in the context of its speakers, their culture, and their needs. ---\citet{Lent2022Jun}
% \end{quote}

\begin{quote}
    If the fleet-footed Achilles and a slow-moving tortoise are in a race, Achilles will never catch the tortoise if the tortoise has a head start. Regardless of how fast Achilles runs, he first has to reach a point the tortoise already passed, by which point the tortoise will have moved ahead. --Zeno's Achilles Paradox \footnote{https://www.britannica.com/topic/Achilles-paradox}
\end{quote}

The majority of research in the NLP community has focused on only a handful of the world's languages~\cite{Joshi2020Jul, Bird2022May}. Particularly, languages spoken by communities in the Global South have largely been neglected~\cite{Nekoto2020Nov, Schwartz2022May}. Languages understudied by the NLP community are usually referred to as `low-resource', while those well-studied are referred to as `high-resource.' This framing of high vs low-resource languages resembles Zeno's Achilles paradox: `high-resourced languages' are the tortoise, that have been given a head start in the research community and continue to receive much of the attention, and `low-resource languages' are Achilles. In reality, Achilles can always outrun the tortoise\footnote{https://ibmathsresources.com/2018/11/30/zenos-paradox-achilles-and-the-tortoise-2/}. However, the face value interpretation of the paradox can serve as an analogy for how the current trajectory of the NLP research community to include majority of the worlds languages in the path already forged for `high-resourced' languages leaves `low-resource languages' constantly trying to catch up to a goalpost that is always moving.

% \citep{} show that the majority of African languages are left behind in NLP studies, while \citet{} argue the colonial legacies of language impact how Indigenous languages are excluded from mainstream NLP research. 
% The lack of representation has also led to implicit defaults: for instance, it is common for papers that work exclusively on the English language not to explicitly name it~\cite{Bender2021Dec}. While in recent years, we have seen an increase in the diversity of languages studied, there is still a gap both in representation and performance~\cite{NigatuTonja2023Dec, Ojo2023Nov}. 

% why is it interesting and important? 
The disparity in research and performance of language technologies across languages can be a double-edged sword. On the one hand, under-studied and underserved languages may be at a higher risk of language loss and have speakers exposed to direct downstream harm due to failures of language technologies~\cite{10.1145/3630106.3658546, Choudhury2023Nov}. On the other hand, the drive to include these languages in research without proper consideration of community needs (1) may lead to aggressive--and at times exploitative--data collection and (2) result in technologies that do not meet the needs of the communities who speak those languages~\cite{10.1145/3530190.3534792, LeFerrand2022May, Dearden2021Mar}.

% why is it hard? 
% Recent works~\cite{} have put in efforts to include historically excluded languages in tasks like Language Modeling~\cite{}, Machine Translation~\cite{}, and Named Entity Recognition~\cite{} to name a few. 

Recently, we have seen efforts to increase the representation of `low-resource languages' in NLP research~\cite[e.g.][]{NLLB2024Jun, adelani-etal-2022-thousand}. Yet, the exact definition of the term `low-resource' remains elusive\footnote{`Under-resource' is a term used interchangeably--and perhaps equally as ambiguously--with `low-resource.'  For brevity, we mainly use the phrase `low-resource' in this paper.}. A common criterion to connote languages as `low' vs `high' resourced is data. 
% For instance, \citet{joshi_state_2020} provide a 6-category taxonomy for classifying languages based on the availability of data. 
However, using data as the only criterion oversimplifies the context of the language itself. Languages dubbed as `low-resource' may vary depending on factors like their number of speakers, non-digital archives, or language experts~\cite{NRCCanada2024Jun}. 

The lack of consensus in what qualifies a language as `low-resource' makes it challenging to \textbf{(1)} track progress in research and development for `low-resource languages' in general, \textbf{(2)}  determine what interventions are effected towards a language, \textbf{(3)} pinpoint when a language stops being `low-resource', and \textbf{(4)} discern if technologies built for these languages truly address the needs of the communities who speak them or if they are built simply on the premise that the same technology exists for a `higher resourced language.'

% what are the key components of our approach and our main results? 
In this work, we survey papers that study languages coined as `low-resource'. We qualitatively analyzed 150 papers that include the keywords `low-resource' and `under-resource.' 
% from the ACL Anthology and popular speech processing conferences: INTERSPEECH and International Conference on Acoustics, Speech, and Signal Processing (ICASSP). 
% searched for the keywords `low-resource' and `under-resource' in the ACL anthology and popular speech processing conferences and randomly sampled 150 papers for manual analysis. 
We used qualitative methods to unravel (1) how such papers define the term `low-resource' or `under-resource', (2) what languages are studied as `low-resource', and (3) what criteria is used to classify a language as `low-resource.' 
% We find that the term is overloaded; with some papers using it to describe `low-resource domains,' i.e. domains or tasks with limited data available, and others using it to describe `simulated low-resource settings', usually involving under-sampling data from a higher-resourced language to demonstrate if a method would work for a language with limited data. We exclude such papers from our analysis and focus on papers that exclusively study `low-resource languages.' 

Our analysis reveals four separate but interacting aspects of `resourcedness' that are used to connote a language as `low-resource' (see Section \ref{definations} \& Section \ref{languages}).
% , we highlight the different languages studied as `low-resource' in our sample. 
In Section \ref{recommendations}, we use real-world examples to demonstrate how each of the aspects interact and how those interactions impact what interventions are designed and implemented for a language. Finally, we use our analysis to ground recommendations for different stakeholders (see Section \ref{sec:recs}).

\section{Methodology}

\paragraph{Data} We collected data for papers published at *CL venues\footnote{We focused on the top 6 venues based on Google Scholar metrics for computational linguistics( \url{https://scholar.google.com/citations?view_op=top_venues&hl=en&vq=eng_computationallinguistics)}} from the ACL Anthology\footnote{https://github.com/acl-org/acl-anthology} and at the following Speech Processing conferences: INTERSPEECH and International Conference on Acoustics, Speech, and Signal Processing (ICASSP) using the Semantic Scholar \cite{kinney2023semantic} API
% \footnote{https://www.semanticscholar.org/}
. We used a keyword search to identify papers that include the terms `low-resource' or `under-resource' in their titles or abstracts. Our final corpus included 868 unique papers.

\paragraph{Qualitative Analysis} 

In the initial stage of our analysis,
we found that the term `low-resource' is used to refer to three broad categories: (1) tasks and domains where there is a lack of labeled data, 
% (e.g. Dialogue \cite{sun_multimodal_2022, feng_schema-guided_2023, sun_multimodal_2022} tasks and medical (e.g. \cite{mrini_gradually_2021} domains.)
 (2) `simulated low-resource' settings via methods like under-sampling
 %to test the performance of a proposed method in `low-resource' settings
, (3) `low-resource languages' defined based on diverse criteria. Table \ref{tab:typesofpapers} summarizes this finding. 
% Both \citet{mager-etal-2020-tackling} and \citet{Zevallos2023Jul} found that the performance improvement in the simulated low-resource setting did not translate to improvement in the actual low-resource languages. 
For our qualitative analysis, we exclusively focused on the third category, i.e., papers that study `low-resource languages' as our interest is in understanding how a language is labeled as low-resource. We also found papers that tried both sampling higher-resourced languages and using actual, low-resourced languages \cite[e.g.][]{Zevallos2023Jul}. We include those in our analysis as they study a `low-resourced language' in addition to a simulated setting.
% we observed some papers used the term to describe a low-resource task or domain (e,g, \citet{Dai2020Jul}) and some papers used simulated low-resource settings--sampling a small amount of data from a high-resource language (e.g. \citet{Dehouck2020Dec}) to show the applicability of their models or methods for low-resource languages.
We manually labeled 541 papers to identify those that explicitly work on non-simulated low-resource languages and randomly sampled 150 papers for qualitative analysis. Our sampling strategy was independent of any parameter such as publication year; the time span for the 150 papers was 2017-2023. We conducted our analysis by reading each paper and annotating how the term `low-resource' or `under-resource' is defined, what languages are studied in the paper, and any additional challenges mentioned in the paper in relation to the languages of study being `low-resource.' 
% We started our analysis with a set of predefined questions (see Appendix \ref{}) and refined them as we progressed through our analysis. 
% We observed that some papers use the term `low-resource' to refer to a domain or a task, not necessarily a language.  ... We labeled the papers based on what  
% We specifically looked at how the papers define the term ``low-resource'' or ``under-resource'', what criteria are used to classify a language as ``low-resource'', and what challenges are mentioned in the paper about working with ``low-resource'' languages. 
% \hellina{is this too HCI?} 
We used inductive thematic analysis~\cite{Braun2006Jan} and discussed the themes that emerged from our analysis in frequent meetings to synthesize overarching themes. In the following section, we present the results of our analysis along with illustrative quotes. 
% inform the design of our experiment to extract information from the full dataset. 
% \hellina{should this be called a pilot? should it be excluded? should it be framed as a qualitative study? }

% \paragraph{Automatic Data Extraction with GPT-4} Based on the results of our manual analysis, we designed a prompt to automatically extract information from the X papers in our dataset. \TODO{Add information on GPT params used.} We first extracted information from 50 randomly sampled papers and manually verified the responses. We then iteratively adjusted our prompt based on the ambiguities we observed in the responses. Appendix \ref{} presents our final prompt. We then used \TODO{add methods here if we have additional analysis} LDA\cite{} topic modeling to uncover common topics in the definitions of the term ``low-resource.''
\begin{table}[h!]
\small
    \centering
    \begin{tabular}{p{0.08\textwidth}|p{0.13\textwidth}|p{0.1\textwidth}|p{0.04\textwidth}}
    \hline
      \textbf{Category}   & \textbf{Description} & \textbf{Examples}  & \textbf{\%} \\
      \hline
     Tasks and Domains & tasks and domains where there is limited labeled data  &  \citet{sun2022multimodal, bajaj2021long} & 27.27 \\
     \hline
     Simulated & using techniques like under-sampling to simulate low-resource settings & \citet{Zevallos2023Jul, Dehouck2020Dec} & 12.27 \\ \hline
     Languages & languages categorized based on factors like data or number of speakers & \citet{coto-solano-2022-evaluating, ponti-etal-2021-parameter} &65.04  \\
     \hline
    \end{tabular}
    \caption{\textbf{Three categories of papers returned for the keyword search for `low-resource.'} Note that the percentages do not add up to 100 because some papers fall into more than one category. For instance, \citet{mager-etal-2020-tackling} study both simulated and actual low-resource languages.}
    \label{tab:typesofpapers}
\end{table}
\section{What is a `low-resource' language?} \label{definations}
In this section, we present the overarching aspects we found from our thematic analysis. It is first important to note the different styles papers use when defining the term `low-resource': 
% With regards to papers depth of definitions, we give the following examples:
% /Some papers state their definitions explicitly, while others cite prior work or allude to what they are referring to as low-resource in the challenges they describe: 
% We found that papers may explicitly state their definition \cite[e.g][]{goyal_flores-101_2021}; allude to the definition in the challenges they describe with working on a low-resourced language \cite[e.g][]{ferger-2020-processing}; cite prior work that defines the term \cite[e.g][]{sun-xiong-2022-language}; state some languages as examples \cite[e.g][]{lane-bird-2020-interactive}; or a combination of any of the above \cite[e.g[][]{lane-bird-2021-local}. 
 
% \subsection{The Terminology is Overloaded} \label{strata}
\begin{quote}
    \textit{``Languages facing this lack of large amount of data are called low-resourced, and all linguistic varieties in Mexico are struggling with this situation.''} --\citet{sierra-martinez-etal-2020-cplm}
\end{quote}

% \begin{quote}
%     Being a low-resource language in terms of standard linguistic resources, recent text simplification approaches that rely on manually crafted simplified corpora or lexicons such as WordNet are not applicable to Urdu. --\citet{}
% \end{quote}

\begin{quote}
    \textit{``Under-resourced, under-studied and endangered or small languages yield problems for automatic processing and exploiting because of the small amount of available data as well as the missing or sparse description of the languages.''} --\citet{ferger-2020-processing}
\end{quote}

\begin{quote}
    \textit{``It frames these as “low resource languages,” lacking the text, speech and lexical resources that are needed for creating speech and language technologies (Krauwer, 2003)''}. --\citet{lane-bird-2021-local}
\end{quote}

In the quotes shown above we see that \citet{sierra-martinez-etal-2020-cplm} explicitly define the term, \citet{ferger-2020-processing} describes challenges of working with low-resource and \citet{lane-bird-2021-local} define the term and provide citations from prior work. If a paper uses prior work without explicitly stating its definition, we rely on the definition of the cited work. In cases where there are no explicit definitions, we rely on the challenges mentioned by the paper to categorize how the paper decides if a language is `low-resource.' 
% Regardless of if it is defined explicitly, w

\begin{figure}
    \centering
\includegraphics[width=0.45\textwidth]{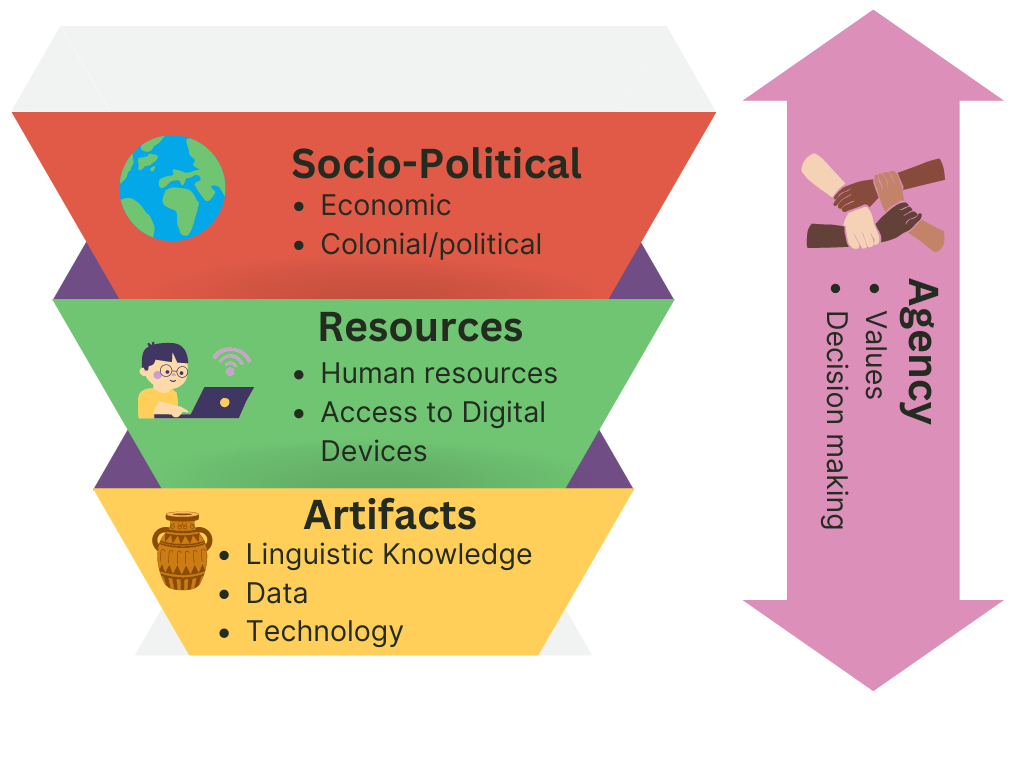}
    \caption{ \textbf{Four overarching aspects that contribute to a language being classified as low-resource.} Socio-political aspects are at the top, influencing both the availability of resources and the creation of artifacts. Community agency is a common thread in all the other three aspects.}
    \label{fig:strata}
\end{figure}

We found that definitions for the term `low-resource' borrow from four aspects: (1) \textbf{Socio-political} aspects relating to financial and historical constraints, (2) \textbf{Resources}, both human and digital, (3) \textbf{Artifacts} such as linguistic knowledge, data, and technological infrastructure, and (4) \textbf{Agency} of community members in what technology is built for their languages. 
We summarize these four aspects in Figure \ref{fig:strata} and dive into detail about each aspect in the following subsections.

% From papers that exclusively study low-resource languages, we synthesized a strata of definitions commonly used in the literature. In Figure \ref{fig:strata}, we show the different levels and how they interact with each other along with example papers. In this section, we discuss each of the levels we observed.

\subsection{Socio-Political}
Some papers call out structural issues pertaining to societal, economic, and political forces. 
%\paragraph{} 
We found papers that reflect on low-resourcedness due to financial and economic constraints to curating data \cite[e.g.][]{coto-solano-2022-evaluating, pathak-etal-2022-asner} and limited use of such languages in mainstream media, government, and education \cite[e.g.][]{mehta-etal-2020-learnings}.  For example:
\begin{quote}
   \textit{``In many of these communities, languages like English and Spanish have displaced the Indigenous languages in domains such as technology and chatting, and so the available data is curtailed.''}--\citet{feldman-coto-solano-2020-neural}
\end{quote}

\begin{quote}
\textit{``However, these languages are not represented in education, government, public services, and media, and therefore, they show high levels of endangerment.''}--\citet{sierra-martinez-etal-2020-cplm}
\end{quote}
% \begin{quote}
%     \textit{``
%     Such modifications, however, cannot address the biggest flaw of our proposed task: it does not respond directly to people’s agenda in terms of language work, but simply tries to leverage people’s skills to respond to westerners’ expectations.''}--\citet{LeFerrand2022May}
% \end{quote}
% Lack of financial resources, which can itself be linked to colonial legacies, may constrain what digital equipment is available to communities and whether they have the resources needed to train linguistic experts. Colonial forces that left their burning legacies on colonized communities resulted in a limited number of human resources, including a reduced number of native speakers\cite{Coffey2021Apr, Birhane2020Aug, Choudhury2023Nov}. 

\subsection{Resource} \label{sec:resources}
The second aspect discussed by papers is the availability of and access to human and digital resources\footnote{Note that in the context of this work, data is an artifact curated for NLP purposes and so is not referred to as a resource in this category.}. 

\paragraph{Human Resources} We found three types of human resources mentioned in papers in relation to low-resource languages: (1) native speakers \cite[e.g.][]{feldman-coto-solano-2020-neural, leong-etal-2022-bloom}, (2) linguistic experts \cite[e.g.][]{pathak-etal-2022-asner}, and (3) NLP researchers \cite[e.g.][]{yimam-etal-2020-exploring}. With regards to native speakers, while some low-resource languages are described as having a limited number of native speakers, others are described as still being low-resourced despite a large number of native speakers. For instance:  
% In addition to the number of native speakers, papers also use the lack of language experts and NLP researchers as reasons to classify a language as `low-resource' (e.g. \citet{yimam-etal-2020-exploring}).
% We observed the number of native speakers being one of the reasons reflected by papers that study `low-resource languages.'

\begin{quote}
    \textit{``Quechua, a low-resource language from South America, is a language spoken by millions but, despite several efforts in the past, still lacks the resources necessary to build high-performance computational systems.''}--\citet{melgarejo-etal-2022-wordnet}
\end{quote}

\begin{quote}
   \textit{``However, low-resource languages such as Amharic have received less attention due to several reasons such as
lack of well-annotated datasets, unavailability of computing resources, and fewer or no expert researchers in the area.''}--\citet{yimam-etal-2020-exploring}
\end{quote}
% \begin{quote}
%    \textit{``Although the term `low resource' is used to describe a wide swath of languages, most Indigenous languages in Canada would be considered `low-resource' in multiple senses of the word, having both a low amount of available data (annotated or unannotated), and a relatively low number of speakers.''}---\citet{Pine2022May}
% \end{quote}

\paragraph{Access to Digital Devices and Platforms}
Lack of access to digital devices--and by extension, the digital presence of communities--is another reason mentioned in relation to `low-resource' languages \cite[e.g.][]{bamutura-etal-2020-towards, nzeyimana-niyongabo-rubungo-2022-kinyabert}. Mainly, this reason is tied to the lack of available digital data for languages that fit the mainstream way of training models. Papers state that `low-resource' languages are not available in formats suitable for crawls and scraping \cite[e.g.][]{feldman-coto-solano-2020-neural}. 
% and that there is limited `online presence' of communities (e.g. \citet{bamutura-etal-2020-towards, nzeyimana-niyongabo-rubungo-2022-kinyabert}).
% Some papers note the lack of digitized data, even when non-digital data is available \cite{}.

\begin{quote}
    \textit{``The included low-resource languages are
also very limited because they are mainly sourced
from Wikipedia articles, where languages with
few articles like Kinyarwanda are often left behind.''} --\citet{nzeyimana-niyongabo-rubungo-2022-kinyabert}
\end{quote}

% \begin{quote}
%     Furthermore, there may be low confidence in using computers and text editors...
%     % , and inadequate support for the language in terms of keyboarding and spelling correction. 
% --\citet{LeFerrand2022May}
% \end{quote}

\begin{quote}
    \textit{``In addition
to this, many Indigenous communities have chronic digital inequalities, which makes it difficult to generate crowd-sourcing campaigns for those languages. Finally, in many cases, the data that is most valuable
to speakers of the language is collected from elders and knowledge keepers, but those elders might
be the people who have the least access to technological means of communication.''} --\citet{feldman-coto-solano-2020-neural}
\end{quote}

% \begin{quote}
%     There is also a severe dearth of online content in Gondi, resulting in members of the tribe having to learn a mainstream language in order to enjoy the benefits of internet connectivity. --\citet{mehta-etal-2020-learnings}
% \end{quote}

\subsection{Artifacts}
The third aspect of resourcedness is tied to the production and accessibility of artifacts: linguistic knowledge, data, and technology.

\paragraph{Linguistic Features and Descriptions}
Papers state how there are limited available linguistic descriptions for `low-resource' languages  \cite[e.g.][]{ferger-2020-processing, sikasote-anastasopoulos-2022-bembaspeech}. 
% Linguistic features, such as  \citet{de-lhoneux-etal-2022-zero} state the lack of ``typological features'' for `low-resource languages.'
Often, linguistic features--such as morphological complexity and typology--are used as reasons why it is difficult to blindly adopt methods that work for high-resource languages, even in cases where there is an equal number of training data \cite[e.g.][]{de-lhoneux-etal-2022-zero}. Standardization--or lack thereof--is another feature mentioned in relation to `low-resourcedness' of languages. Both linguistic features and lack of standardization are mentioned as reasons for data sparsity. For example:

\begin{quote}
    \textit{``Due to differences in language typology, it is not necessarily as simple as looking only at number of lines of training data.[...] For example, Inuktitut is known to be highly morphologically complex, resulting in many words (defined as space/punctuation separated) that appear just once or only a few times, even in such a large corpus.''}--\citet{knowles-littell-2022-translation}
\end{quote}

\begin{quote}
     \textit{``Not only is data scarce, but it might lack standardization, making the dataset more sparse than it would be for languages with standardized orthographies and numerous speakers.''} --\citet{coto-solano-2022-evaluating}
\end{quote}

\paragraph{Data}
With regards to data, the classification of a language as low-resource could be based on labeled or annotated data \cite[e.g.][]{ponti-etal-2021-parameter}, unlabeled data \cite[e.g.][]{imanigooghari-etal-2022-graph}, or benchmark data \cite[e.g.][]{reid-etal-2021-afromt}. Some papers focus their definitions on the quality of data \cite[e.g.][]{maillard-etal-2023-small, ramnath-etal-2021-hintedbt}, stating that low-resource language data is usually noisy. Other papers quantify the amount of data \cite[e.g.][]{biswas-etal-2020-semi-supervised}. We also observed a subset of papers that use a predefined cutoff for the amount of data: for instance, \citet{ramachandran-de-melo-2020-cross} state they \textit{``...picked six languages that had around 10K or fewer verses available.''} Some papers would quantify the amount of data in relation to a popular trend in the field:

\begin{quote}
    \textit{``Only some of the 22 scheduled Indian languages, which are a subset of the numerous languages spoken and written in India, have enough resources for training a deep learning model.''} --\citet{saurav-etal-2020-analysing}
\end{quote}

\paragraph{Technology}
Exclusion from technological advances for the languages of study is another aspect mentioned in relation to low-resource languages. This ranges from the lack of basic computational tools--such as text pre-processing tools \cite[e.g.][]{niyongabo-etal-2020-kinnews}
% , pine-etal-2022-requirements}
--to exclusion from pre-trained language models \cite[e.g.][]{leong-etal-2022-bloom, pfeiffer-etal-2020-mad}. 
There were also mentions of lack of compute resources \cite[e.g.][]{yimam-etal-2020-exploring}. 

% With inclusion in large models, reasons could be as specific as `having a writing script not seen in the model'\cite{preiffer} and `languages not included in XLM-R model'\cite{}, or more general `languages not included in popular pre-trained models'\cite{}.

\begin{quote}
    \textit{``Handling utterances with non-Kanien’kéha characters would have required grapheme-to-phoneme prediction capable of dealing with multilingual text and code-switching, which we did not have available.''} --\citet{pine-etal-2022-requirements}
\end{quote}

\begin{quote}
    \textit{``In total, we can discern four
categories in our language set: 1) high-resource languages and 2) low-resource languages covered by the pretrained SOTA multilingual models (i.e., by mBERT and XLM-R); as well as 3) low-resource languages and 4) truly low-resource languages not covered by the multilingual models''}--\citet{pfeiffer-etal-2020-mad} 
\end{quote}

\subsection{Agency}
Transcending all the other aspects is community agency and the role it plays in what and by whom language technologies are built. \citet{coto-solano-2022-evaluating} state how even in cases where communities are willing to provide data, financial constraints prevent them from doing so. \citet{LeFerrand2022May} emphasize building language tools detached from community practices leads to technologies with minimal utility to the communities. This detachment from community practices is also stated as a reason for minimal studies in these languages:
% a point further elaborated for the Assamese language below:
\begin{quote}
     \textit{``Although Assamese has a very old and rich literary history, technology development in NLP is still in a nascent stage.''} --\citet{pathak-etal-2022-asner}
\end{quote}

When communities are actively engaged, we observe their values embedded in the production of technology, regardless of the outcome of the research project:

\begin{quote}
   \textit{``While a total of 24 hours of audio were recorded, members of the Kanien’kéha-speaking community told us it would be inappropriate to use the voices of speakers who had passed away, leaving only recordings of Satewas’s voice. [...] The resulting speech corpus comprised 3.46 hours of speech.''} --\citet{Pine2022May}
\end{quote}

% % \paragraph{\textbf{Recommendations}}
% \begin{figure}
%     \centering
%     \includegraphics[width=0.5\textwidth]{latex/images/strata.png}
%     \caption{Four overarching aspects that contribute to a language being classified as low-resource. Socio-political aspects are at the top, influencing both the availability of resources and the creation of artifacts. Community agency is a common thread in all the other three aspects.}
%     \label{fig:strata}
% \end{figure}
% \section{Low-resource language definitions in NLP papers}

% \begin{quote}
%     On the other hand, data annotation for a language remains a time consuming and expensive process. It is a major challenge for many resource-poor languages, such as Assamese (Glottocode: assa1263), as it requires language experts to perform annotation tasks on a large amount of data. --\citet{pathak-etal-2022-asner}
% \end{quote}

\section{What Languages are Studied as `Low-Resource'?} \label{languages}
% In this section, we describe languages studied as low-resource languages in our data. 
% \begin{figure}
%     \centering
%     \includegraphics[width=\linewidth]{latex/images/pp2.png}
%     \caption{Caption}
%     \label{fig:top_40}
% \end{figure}

% \begin{figure}
%     \centering
%     \includegraphics[width=\linewidth]{latex/images/top_40_langs.png}
%     \caption{Top-40 most studied languages}
%     \label{fig:top_40}
% \end{figure}
\begin{figure}
    \centering
    \includegraphics[width=\linewidth]{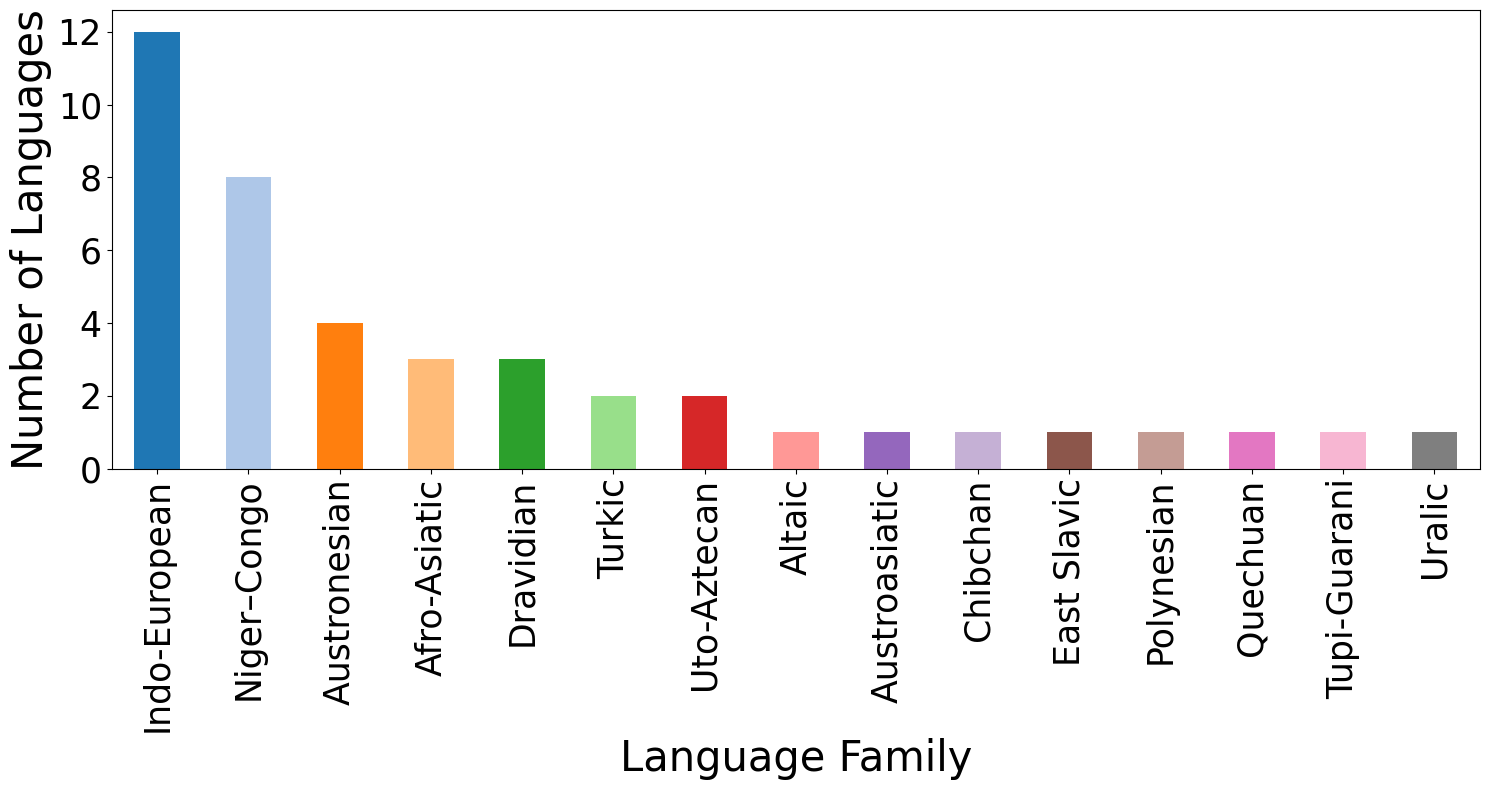}
    \caption{Number of languages included in the studies per language family.}
    \label{fig:lang_fam}
\end{figure}
% Figure \ref{} shows the top 40 languages and their criteria for being categorized as low-resource. As we discussed in Sections \ref{} and \ref{}, multiple factors are used to categorize one language as high or low-resourced. Figure \ref{} demonstrates criteria used in the papers to categorize the top 40 languages shown in the figure as low-resource. 

% Niger-Congo, Quechuan, Uto-Aztecan, Polynesian, Tupi-Guarani, Indo-European, Afro-Asiatic,  East Slavic, Dravidian, Turkic, Altaic, Chibchan, Austroasiatic, Uralic, Austronesian. 
% While not shown in the top 40, there were also languages with only one paper working on them from our sample: for example, there was only one paper, \citet{} working on \TODO{Add language with one paper working on it}. 

Languages may be studied in multilingual contexts, i.e. included alongside other languages \cite[e.g.][]{adelani-etal-2022-thousand, goyal2022flores} or in monolingual contexts \cite[e.g.][]{yimam-etal-2020-exploring, pathak-etal-2022-asner}. Papers had varying depths of descriptions for the languages they studied, with papers working on fewer languages having more in-depth descriptions. For instance, \citet{mehta-etal-2020-learnings}, which exclusively work on the Gondi language, has a dedicated section on the historical, political, and linguistic context of the Gondi language and its community. On the other hand, \citet{goyal2022flores}, which works on 101 languages, has one table with all the languages, their ISO codes, language families, writing scripts, and the amount of available data. 

In Figure \ref{fig:lang_fam}, we show the number of languages and language families studied in our samples,
% the top 20 most frequently studied languages in our sample, 
where papers explicitly mention them as low-resource. 
% We see that Swahili and Telugu take the lead with 14 papers working on them. 
We observe a diverse set of language families, with Indo-European languages having the highest number of languages studied in our samples, followed by Niger-Congo and Austronesian. In Appendix \ref{apn:studed-langs}, we detail the top 20 most frequently studied languages in our sample.

\begin{figure}
     \centering
     \includegraphics[width=0.5\textwidth]{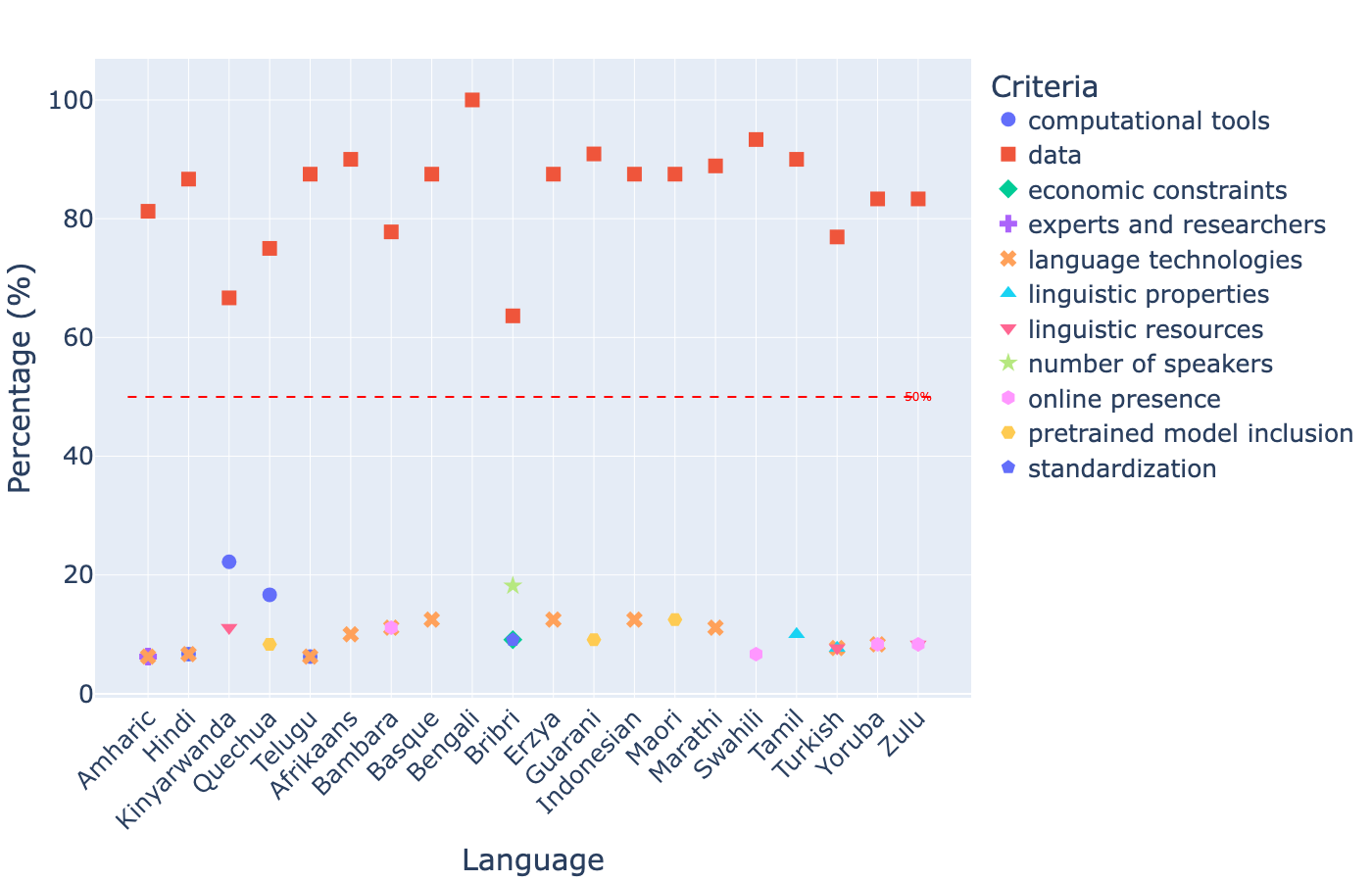}
     \caption{Criteria distribution used in the top-20 languages to categorize languages.}
     \label{fig:catagories}
 \end{figure}

The graph in Figure \ref{fig:catagories} shows the distributions of the various criteria used for categorizing a language as `low-resource' in the top 20 languages studied. While data is the most commonly used criterion across many papers and languages, other factors, such as lack of computational tools, limited number of native speakers, etc, are also used (see Section \ref{definations}). Even with papers that use data as a criterion, we observe different qualifications for \textit{what type} of data a language may lack to qualify as a `low-resource' language.  In Figure \ref{fig:data_cate}, we further break down the criterion of data. We observe that lack of labeled data is the most commonly used criterion in our sample at 39.8\%. We also observe the lack of digitized text (1.7\%) and online-available data (6.9\%) as criteria to connote a language as low-resource. 
% even further. 
% As this is a qualitative study, it is important to note that we cannot make quantitative conclusions on which 

% \TODO{Add figure here for which each language and what criteria were used to label it as low resource}

% \TODO{Analysis on the same language having different categories based on different criteria}

% \section{Discussion}
% \subsection{Language Profiles}
% \hellina{can we have an example with one low-resourced language showing how each of the definitions relates to it?}

% \TODO{broad description of strata and causal links among them}

% \TODO{languages the boundaries are not clear, a reflection on the language analysis.}

% why we should care about the definations and what should we do about it?

\section{Why does it matter?} \label{recommendations}
\begin{figure*} [th!]
     \centering
    \includegraphics[width=\linewidth]{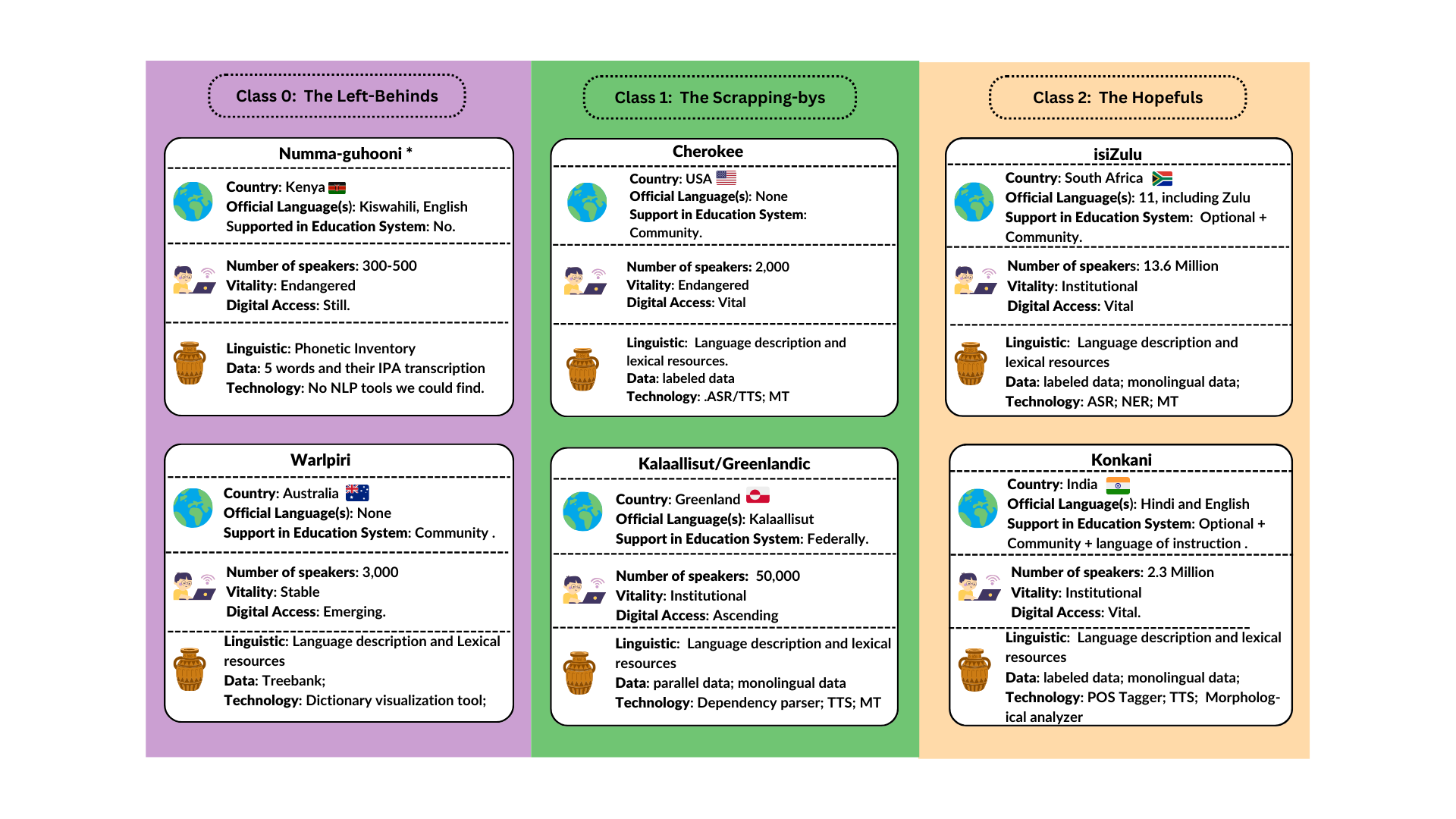}
    \caption{\textbf{Language profiles for six languages across three classes based on data availability.} The first row in each profile deals with socio-political issues, the second row resources, and the last row with artifacts (see Figure \ref{fig:strata}). We observe drastic differences between languages of the same class. See Appendix \ref{apn:labels} for details on the labels.}
    \label{fig:language_profiles}
\end{figure*}

% Thus far, we have highlighted the lack of consensus on the term `low-resource.'
% and described the separate but interacting aspects used to categorize a language as a `low-resource' language.
In the previous section, we describe four overarching aspects that determine if a language is `low-resource': socio-political aspects, human and digital resources, artifacts, and agency of community members.  In Figure \ref{fig:language_profiles}, we present language profiles for 6 languages. We choose the six languages from the bottom three classes in \citet{Joshi2020Jul}: `The Left Behinds' with limited labeled and unlabeled data, `The Scraping-Bys' with some amount of unlabeled data, and `The Hopefuls' with some labeled data. We use literature about these languages and their communities to demonstrate why it matters that we are specific in the terminology we use.
% and how, with in the same class of data availability, there is variance in other resources available for a language. 

% We start by presenting language profiles for 6 languages, 2 from each of the bottom three classes in \citet{joshi_state_2020} and highlighting the differences within and across classes. We then show links and interactions between the different aspects described in Section \ref{definations}. Finally, we argue how the \textit{specific resource} the term `low-resource' refers impacts what is allocated for a particular language and give recommendations for terminology to use to make this more explicit. 

\paragraph{Languages in the same class of data availability might differ in other aspects.} 
From `The Left Behinds', we present profiles for Numma-guhooni\footnote{While this language is refereed to with another name in the literature, there is evidence that the word is derogatory and so we exclusively use the name native speakers use~\cite{stiles_1982}. } and Warlpiri. Numma-guhooni is spoken in Kenya where the official Federal languages are Kiswahili\footnote{also known as Swahili in English speaking contexts.} and English. Warlpiri is spoken by the Warlpiri people of Australia, where the most dominant language is English.
While both languages fall into the same class, the number of speakers for Warlpiri is 4 times that of Numma-guhooni. Ethnologue classifies Warlpiri as a \textit{stable} language, while Numma-guhooni is \textit{endangered}. In terms of digital resource availability, Ethnologue classifies Numma-guhooni as \textit{still} meaning, there is no sign of digital support for the language, while Warlpiri is labeled \textit{emerging} with some digital content available. Warlpiri also has some NLP tools available, for instance, KirrKirr is a dictionary visualization tool for the Warlpiri language~\cite{Manning2001Jun}.

From `The Scraping-Bys', we look at Cherokee and Kalaallisut. Cherokee, spoken by around 2,000 out of the 300,000 Cherokee people of the Cherokee Nation in the United States of America, is labeled as \textit{endangered} by Ethnologue. On the other hand, Kalaallisut, which is spoken by about 50,000 people and is the official Federal language of Greenland, is labeled as \textit{institutional} by Ethnoluge. However, Cherokee has a higher ranking for digital language support, dubbed \textit{vital} while Kalaallisut is \textit{ascending}.
% As a Federal language, Kalaallisut is used in the education system of Greenland. On the other hand, Cherokee is supported by community efforts, including language programs by the Cherokee Nation Language Department\footnote{https://language.cherokee.org/} and the Cherokee Immersion School\footnote{https://www.cwyschools.org/}. 
% While there are artifacts available for both Cherokee and Kalaallisut, the former is mainly sustained though community efforts while the latter is supported federally. 
% As the result,  

% Cherokee language is \textit{endangered} with \textit{vital} digital language support. As the federal language, Kalaallisut is used in the education system of Greenland and labeled as \textit{institutional} by Ethnoluge, with a digital access category of \textit{ascending}, meaning there are some localized tools and machine translation systems.

% While there are no official federal languages in the US, English is the most dominant language, followed by Spanish. Federal polices in the US up to 1948 forced Indigenous children to assimilate to Western culture, punishing students for speaking their languages\cite{jessica2024Jun}. 

% Efforts to support the language in the education system include language programs by the Cherokee Nation Language Department\footnote{https://language.cherokee.org/} and the Cherokee Immersion School\footnote{https://www.cwyschools.org/}. In terms of digitally access, Ethnologue labeled Cherokee as \textit{vital}, meaning it has localized tools as well as machine translation and speech processing tools. 

For `The Hopefuls', we look at isiZulu and Konkani. We observe the two languages are somewhat similar in terms of human and digital resources, with both being \textit{institutional} in vitality and \textit{vital} in digital access.  However, we see the languages vary by their number of speakers with isiZulu having about 6 times the number of speakers as Konkani. Additionally, isiZulu is the most common language spoken as a first language in South Africa, while Konkani has shown a decline in number of speakers, with speakers outside of its primary province declaring other, dominant languages as their native language~\cite{RAJAN2020100299}. Both languages have NLP tools available for tasks like machine translation and speech processing as well as pre-processing tools.
% (see Appendix \ref{apn:resources})

% Overall, we observe less variance in class 2 as opposed to classes 0 and 1.
Overall, we observe that within a given class based on data availability, there are drastic differences in what other resources are available for a language. We observe that the variance decreases as we move up the classes, which can partially be explained by the stark 88.38\% of the world's languages belonging to `The Left-Behinds', compared to 5.49\% in `The Scraping-Bys' and 0.36\% in `The Hopefuls'~\cite{Joshi2020Jul}. However, as we demonstrate, the realities of each of the languages within each class are very different.
\begin{table*}[th!]
\small
    \centering
    \begin{tabular}{|>{\raggedright}p{0.1\textwidth}|>{\raggedright}p{0.12\textwidth}|p{0.25\textwidth}|p{0.45\textwidth}|}
    \hline
    \textbf{Aspect} & \textbf{Sub-Division} & \textbf{Terminology} & \textbf{Definition}  \\
    \hline
    \multirow{2}{*}{Socio-political} & Economic &
       low-affluence~\cite{survey_hammarstrom} & based on Gross Language Product (GLP) (product of the number of native speakers in any country and the country's per capita Gross National Product.) \\ \cline{2-4}
      & Political  & politically-disadvantaged & languages not used in mainstream media and governmental communications due to political forces \\
    \hline
     \multirow{3}{*}{Resources} & Native Speakers* & extinct; critically endangered; severely endangered; definitively endangered; unsafe; safe \cite{unesco_2003} & 6 point scale based on number of speakers of the language \\ \cline{2-4}
    & Online Presence & Low-Web Resource\cite{patil-etal-2022-overlap} & limited online corpus or web presence\\ \cline{2-4}
    & Language experts & expert-constrained & limited number of linguistic experts or researchers  \\ 
    \hline
    \multirow{3}{*}{Artifacts} & {Linguistic Knowledge*} & oral languages; non-native orthography; native orthography & based on the availability and type of orthography a language has. \\ \cline{3-4}
    & & undocumented; inadequate; fragmentory; fair; good; superlative \cite{unesco_2003} & 6 point scale based on the amount and quality of documentation available for a language. \\ \cline{2-4}
    & Data* & Class 0; Class 1; Class 2; Class 3; Class 4; Class 5 \cite{Joshi2020Jul} & 6 classes based on the availability of labeled and unlabeled data  \\ \cline{2-4}
    & Technology* & Still; Emerging; Ascending; Vital; Thriving \cite{simons-etal-2022-assessing} & 5-level classification based on digital language support available in a given language. \\
    \hline
    \end{tabular}
    \caption{\textbf{Suggestions for explicit terminology addressing three aspects we identified through our analysis.} We provide citations for terminology taken from prior work. (*) indicate the terminology are part of a scale and all labels in the scale are listed.}
    \label{tab:recs}
\end{table*}

\paragraph{The different aspects that determine `low-resourcedness' have causal links.} \label{causal_links}
The four aspects we discuss in Section \ref{definations} interact with each other in constraining what is available. \textbf{Socio-political} issues constrain what \textbf{Resources} are available for a given language, which in turn impact what \textbf{Artifacts} are produced for that language. For instance, while there are no official languages in the USA or Australia, federal policies in the US up to 1948 forced Indigenous children to assimilate into Western culture, punishing students for speaking their languages~\cite{jessica2024Jun}. Similarly, colonization destroyed several languages of Indigenous populations in Australia~\cite{doi:10.1080/13549839.2015.1036414}. As a result, both Cherokee and Warlpiri, along with the numerous other Indigenous languages of the Americas, Australia, and Canada are endangered, i.e lack \textbf{human resources}.  

Assimilation is not limited to the languages of the colonizer. Post-independence from colonial rule of Britain, Kenya adopted the educational and language policies of Britain, with English declared the official language in formal sectors and Kiswahili the national language of the country. As a result, the majority of data available in \textbf{digital and electronic media} as well as in public settings are in English or Kiswahili~\cite{Barasa2023Sep}. Hence, speakers of languages like Numma-guhooni are largely assimilated with larger ethnic groups and Kiswahili is predominantly spoken and learned by the new generation~\cite{dahalo_1992}. While in 2010, the Kenya constitution shifted towards centering the preservation of native languages, there were not enough funds allocated to carry this through~\cite{Barasa2023Sep}. Though at a different scale, this is similar to the case of Konkani, which is in `The Hopefuls' class, losing native speakers to more dominant local languages~\cite{RAJAN2020100299}.

Constraints of human and digital resources restrict the creation of \textbf{artifacts} for languages. As discussed in Section \ref{sec:resources}, the minimal digital presence results in limited \textbf{available data}, especially at the scale needed for training SOTA \textbf{models.} Links among the different aspects are not necessarily linear; socio-political issues also directly constrain what languages are taught in schools, impacting \textbf{linguistic knowledge} produced for a language. Additionally, prior work demonstrates the Western-dominated \textbf{researcher} landscape in NLP and how it ties to coloniality~\cite{Held2023Nov}. With the limited number of speakers for a given language, the number of NLP researchers who are also native speakers of the language is largely constrained, which is further confounded by the limited financial resources available to researchers from such communities. As a result, having \textbf{agency} in what tools are designed for a language becomes challenging.

\paragraph{Knowing which aspect a language is lacking in allows for targeted interventions.} One of the main factors that determine the survival of a language is \textit{inter-generational transmission} \cite{unesco_2003}. 
% Hence, for languages like Cherokee, Numma-guhooni, and Warlpiri, interventions would be more fruitful if targeted towards langauge teaching. 
For instance, while Cherokee and Kalaallisut are both in the same class, Cherokee is \textit{endangered} while \textit{Kalaallisut} is institutional. Hence, interventions--both in socio-political and artifact aspects--are best targeted toward reviving and preserving the Cherokee language. On the other hand, digital access for Kalaalisut is \textit{ascending}, hence there might be more efforts towards increasing the availability of digital data. Since Kalaallisut is institutional, financial resources for preserving and growing the language are available at a federal level. Additionally, it is used as the language of instruction in the education system of the country, aiding in the inter-generational transfer of the language. Across classes, we observe similarities in Numma-guhooni and Konkani, of native speakers assimilating to other dominant but local languages. Hence, interventions for these languages may be more effective in language learning apps that focus on learning the less-dominant language and translation systems between dominant local languages and the target language. 

\textbf{Communities are actively resisting exploitation and sustaining their languages; our tools should support them.} Despite the several layers of constraints, it is important to note that communities are not in idle state of deficit.
Across classes, we observe a similarity between Warlpiri and Cherokee, in that there are community-based initiatives to preserve and grow the languages (e.g. the  Warlpiri Education and Training Trust (WETT)\footnote{https://www.clc.org.au/wett/} and the Cherokee Immersion School\footnote{https://www.cwyschools.org/}). By centering community values in our designs and research, we can collectively forge new paths for each language, conditioned on its unique circumstances.

\section{What can we do?} \label{sec:recs}
\paragraph{Using specific terminology or having explicit definitions allows us to measure progress more precisely.} 
The specific \textit{resource} a language is deemed `low' in directly impacts what interventions are effected towards it. For instance,
% we observe differences in what resources are available for a language affecting whether or not it is included in programs that aim to increase representation of languages in 
programs aimed at increasing language representation in Human Language Technologies (HLTs) have several selection criteria \cite{Cieri2016May}.
% --which may come from government agencies \cite[e.g.][]{LORLEI2023Dec, METANET2010} and industry \cite[e.g.][]{NLLB2024Jun}---
Such programs use different terminologies and definitions,  where ``each term encodes differences in traditions, goals, and approaches'' \cite{simpson2008}. As a result, what languages are included and served by such programs differ, even if languages have the same amount of data.  

While Cherokee is tagged as having \textit{vital} digital resources, it is also an \textit{endangered} language. Collecting more data in the language from the limited number of speakers or including it in Large Language Models may not exactly alleviate its low-resourcedness. 
% Prior works have used terminology to highlight some aspects of resourcedness; for instance 
We argue for more explicit declarations of \textit{which} aspects of resources are being referred to when the term low-resource is used. In Table \ref{tab:recs}, we give recommendations for terminologies based on prior work and our findings. There are also several taxonomies and classes based on data \cite[e.g][]{Joshi2020Jul}, language vitality \cite[e.g][]{unesco_2003},  and digital support \cite[e.g][]{simons-etal-2022-assessing}.

\paragraph{Recommendations for stakeholders:}
Based on our findings, we give recommendations for different stakeholders involved in the effort to increase language representation in NLP research.
% \begin{itemize}
    % \item 
    Individual researchers can (1) engage with community members and speakers of the languages they work on, (2) articulate how their work is limited in relation to the characteristics of the languages they work on, and (3) be explicit about what criteria they use to denote a language as `low-resource.' 
    % \item 
    Community members can also form grassroots organizations such as Masakhane\footnote{https://www.masakhane.io/}, which allow researchers who speak diverse languages to build language technologies together and learn from each other's experiences. Additionally, such organizations can prioritize engaging with native speakers who may not be in the NLP research field,  allowing for diverse perspectives when deciding what tools should be built for what language.
    % language speakers and grassroots research communities
    % \item 
    % Conferences: 
    Workshops such as AmericasNLP\footnote{https://github.com/AmericasNLP} and AfricaNLP\footnote{https://africanlp.masakhane.io/} continue to serve as spaces for fostering research and collaboration for languages that are mostly ignored in mainstream NLP research. However, main (*)CL conferences can increase the representation of these languages by (1) offering alternative tracks for papers, (2) easing the cost of attendance and registration for researchers from these communities, and (3) diversifying conference venues.  
    % alternative tracks for such languages; paywalls in publishing; locations of conferences; workshops
    % \item Academic Institution: 
    Academic institutions can aid researchers who speak these languages by
    % advance low-resource language technology by fostering research, offering specialized education, and scholarships to underrepresented researchers. They can establish research centers, 
    promoting interdisciplinary collaboration and partner with local and international organizations to document and preserve marginalized languages.
    % link between institution to institution collaboration. and institution to the community (language speakers)
    % \item Industry: 
    Industry players interested in language diversity of their products can play a role by offering financial and technical support; for example, subsidizing resources for communities working on low-resource languages. Companies could also prioritize making their products accessible to the communities \cite[e.g.][]{ustun2024aya}.
% providing funding Google; Cohere data collection; 
    % \item Government Bodies: 
    Government bodies can play a role in preserving languages through policies, funding, and digital inclusion. 
    % Collaboration with academia and industry drives innovation while maintaining linguistic diversity in technological advancements.
    % \item Funding Agencies: 
    Funding agencies can support language diversity and enforce building  technologies that are relevant to the specific linguistic community by setting research priorities and prioritizing grants to underrepresented researchers. 

\section{Conclusion}
In this paper, we present 4 aspects of `resourcedness' used to classify a language as `low-resource' based on a qualitative survey of 150 papers. Based on our analysis, we give recommendations for terminology that explicitly calls out which resource we are referring to when we say a language is `low-resource.' A language may lack in several aspects, making the use of individual terminology difficult--e.g. in multilingual settings.
% This difficulty to document at scale has also been reflected in prior work \hellina{find the paper that says it is hard to document as models increase in size, maybe datasheets for datasets}.
However, the difficulty does not absolve us from the responsibility to provide detailed documentation. At the very least, clear statements on what exactly is meant by low-resource when referring to a language would allow us to more clearly articulate the problems a particular technology resolves for a particular language. 
\section{Limitations} 
As a qualitative study, our paper does not give the definitions of the term from all the papers in all the venues we searched. We also do not make quantitative claims. Instead, we focus on a nuanced analysis of how our sample papers describe the phenomenon and provide direct quotes from papers we analyzed as evidence. While it was not practical for us to conduct qualitative analysis on more than the papers in our sample, future work could use automated methods and conduct a quantitative analysis. Similarly, our analysis of what languages are studied is limited to the papers in our sample. This could also be supplemented with automated extraction at scale. Additionally, while we could not perform a longitudinal analysis with our sample size of 150 papers, future work could explore such a study to understand how the use of the term `low-resource' evolved over time. 

\bibliography{anthology,custom}

% \section*{Appendix}
\appendix

\label{sec:appendix}

\section{Labels for Classifying Languages} \label{apn:labels}

In this section, we provide the descriptions for labels used for language vitality and digital access used in Figure \ref{fig:language_profiles}. 
% As well as the scales and taxonomies used in Table \ref{tab:recs}.

\subsection{Vitality}
In this work, we refer to 
% two taxonomies for language vitality. For Figure \ref{fig:language_profiles}, we used 
the scale from Ethnologue\footnote{https://www.ethnologue.com/} which is derived from the Expanded Graded Intergenerational Disruption Scale (EGIDS) \cite{egis}. 

\paragraph{Institutional} — The language has been developed to the point that it is used and sustained by institutions beyond the home and community.
\paragraph{Stable} — The language is not being sustained by formal institutions, but it is still the norm in the home and community that all children learn and use the language.
\paragraph{Endangered} — It is no longer the norm that children learn and use this language.
\paragraph{Extinct} - The language is no longer used, and no one retains a sense of ethnic identity associated with the language.

% For Table \ref{tab:recs}, we also include the scale from \cite{unesco_2003} described below.

% \paragraph{}
\subsection{Digital Access}
This taxonomy is from \citet{{simons-etal-2022-assessing}} and is also used by Ethnologue. 
\paragraph{Still} — this language shows no signs of digital support
\paragraph{Emerging} — the language has some content in digital form and/or encoding tools
\paragraph{Ascending} — the language has some spell checking or localized tools or machine translation as well
\paragraph{Vital} — the language is supported by multiple tools in all of the above categories and as well as some speech processing
\paragraph{Thriving} — the language has all of the above plus virtual assistants
\section{Criteria used in Studying Languages}
\begin{figure*}[h!]
    \centering
    \includegraphics[width=\linewidth]{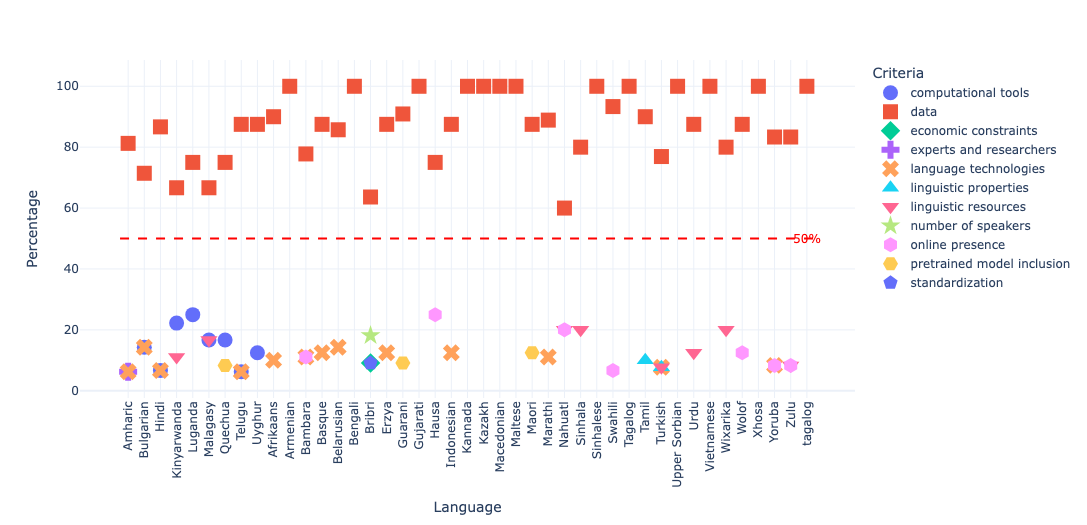}
    \caption{Distribution of criteria stated by papers in our study to categorize languages as low-resource.}
    \label{fig:all_langvscr}
\end{figure*}

\begin{figure}[h!]
     \centering
     \includegraphics[width=0.5\textwidth]{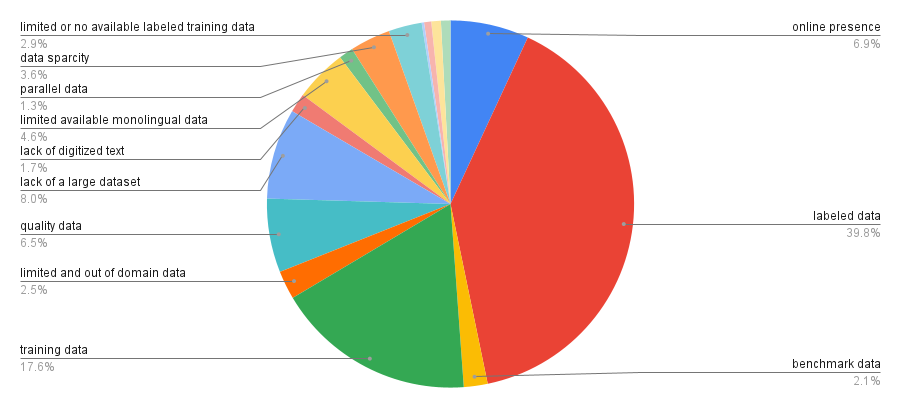}
     \caption{Criteria used in the papers to show lack of data.} 
     \label{fig:data_cate}
 \end{figure}

Figure \ref{fig:all_langvscr} shows the distributions of
the various criteria used for categorizing a language
as ‘low-resource’ in the studied languages.
Figure \ref{fig:data_cate} depicts different perspectives used to refer to the lack of a dataset for a language. 

\section{Most frequently
studied languages} \label{apn:studed-langs}
\begin{figure}[h!]
    \centering
    \includegraphics[width=\linewidth]{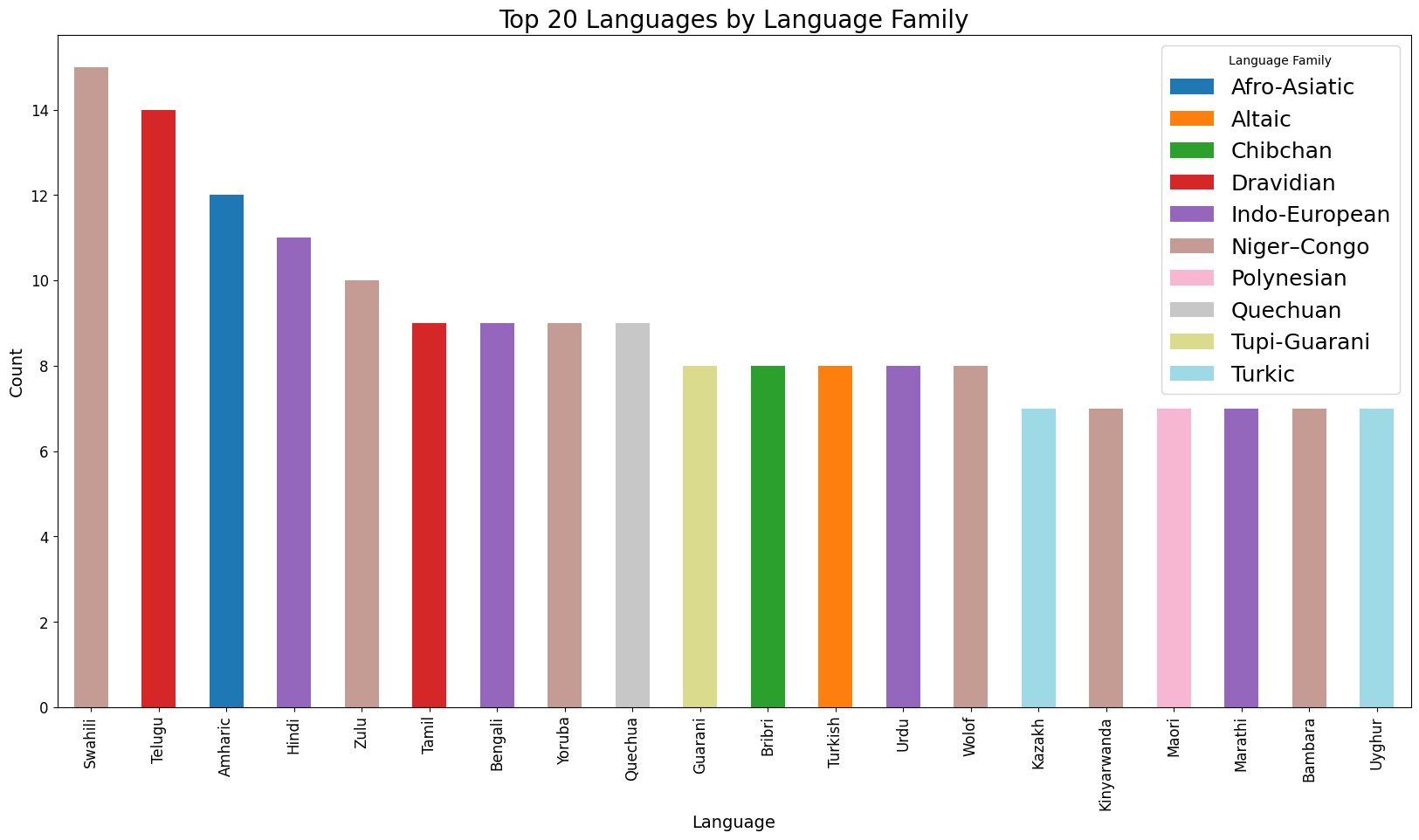}
    \caption{Number of papers per language for the top-20 most studied languages.}
    \label{fig:top_20}
\end{figure}

Figure \ref{fig:top_20} shows the top 20 most frequently studied languages in our sample. We see that Swahili and Telugu take the lead with 14 papers working on them. Geographically, we observe that Indian languages ($n=7$) are the most represented in our sample, with an equal number of languages ($n=7$) from the entire continent of Africa. 
\section{Categories used to define low-resource }
Here, we grouped papers according to the criteria used in the paper to categorize a language as a low-resource language. 
\paragraph{Socio-political}
[{\cite{maillard-etal-2023-small,coto-solano-2022-evaluating,pathak-etal-2022-asner}]

\paragraph{Resources}
\subparagraph{Native Speakers} 
[{\cite{pine-etal-2022-requirements,oliver-etal-2022-inflectional,coto-solano-2022-evaluating,feldman-coto-solano-2020-neural,leong-etal-2022-bloom}]
\subparagraph{Online Presence}
[{\cite{bamutura-etal-2020-towards,sierra-martinez-etal-2020-cplm,adelani-etal-2022-thousand,nzeyimana-niyongabo-rubungo-2022-kinyabert,feldman-coto-solano-2020-neural,bustamante-etal-2020-data,patil-etal-2022-overlap,adelani-etal-2022-thousand}]
\subparagraph{Language experts} 
[{\cite{brixey-etal-2020-exploring,yimam-etal-2020-exploring}]

\paragraph{Artifacts}
\subparagraph{Linguistic Knowledge}
   [{\cite{qasmi-etal-2020-simplifyur,coto-solano-2022-evaluating}]
\subparagraph{Data}
 [{\citet{ferger-2020-processing,zevallos-bel-2023-hints,pine-etal-2022-requirements,fei-li-2020-cross,eskander-etal-2020-morphagram,xia-etal-2021-metaxl,goyal-etal-2022-flores,sorokin-2020-getting,pfeiffer-etal-2020-mad,sierra-martinez-etal-2020-cplm,ahuja-etal-2022-calibration,mehta-etal-2020-learnings,le-ferrand-etal-2022-learning,mukiibi-etal-2022-makerere,chaudhary-etal-2021-reducing,ustun-etal-2020-udapter,eskander-etal-2020-unsupervised,liang-etal-2022-label,pfeiffer-etal-2021-unks,imanigooghari-etal-2022-graph,dione-etal-2022-low,chukwuneke-etal-2022-igbobert,schmidt-etal-2022-dont,hasan-etal-2020-low,muradoglu-hulden-2022-eeny,biswas-etal-2020-semi-supervised,marchisio-etal-2022-bilingual,maillard-etal-2023-small,litschko-etal-2020-towards,coto-solano-2022-evaluating,gaim-etal-2023-question,adebara-etal-2022-linguistically,krishnan-ragavan-2021-morphology,alabi-etal-2020-massive,yimam-etal-2020-exploring,li-etal-2022-multi-level,saunack-etal-2021-low,niyongabo-etal-2020-kinnews,ramnath-etal-2021-hintedbt,ponti-etal-2021-parameter,adouane-etal-2020-identifying,reid-etal-2021-afromt,parovic-etal-2022-bad,minixhofer-etal-2022-wechsel,zeng-etal-2023-soft,pathak-etal-2022-asner,botha-etal-2020-entity,chakrabarty-etal-2022-featurebart,debnath-etal-2021-towards,sarioglu-kayi-etal-2020-detecting,alabi-etal-2022-adapting,ko-etal-2021-adapting,liu-hulden-2020-analogy,wang-etal-2020-structure,zhou-etal-2020-improving-candidate,sharma-etal-2022-hawp,bari-etal-2021-uxla,imanigooghari-etal-2023-glot500,yuan-etal-2020-interactive,gezmu-etal-2022-extended,qi-etal-2022-enhancing,knowles-littell-2022-translation,khayrallah-etal-2020-simulated,mager-etal-2020-tackling,monsur-etal-2022-shonglap,ramachandran-de-melo-2020-cross,sun-xiong-2022-language,hangya-etal-2022-improving,saurav-etal-2020-analysing,ouyang-etal-2021-ernie,parvez-chang-2021-evaluating,moeller-etal-2021-pos,fomicheva-etal-2022-mlqe,mueller-etal-2020-analysis,siddhant-etal-2020-leveraging,bartelds-etal-2023-making,daniel-etal-2019-towards,chen-etal-2022-bridging,fetahu-etal-2022-dynamic,li-etal-2022-low,li-etal-2022-low,bartelds-wieling-2022-quantifying,minixhofer-etal-2022-wechsel,minh-etal-2022-vihealthbert,koloski-etal-2022-thin,coto-solano-etal-2022-development,yakut-kilic-pan-2022-incorporating,linke-etal-2022-conversational,langedijk-etal-2022-meta,muradoglu-hulden-2022-eeny,huang-etal-2022-unifying,jundi-etal-2023-node,xu-etal-2023-unsupervised,li-etal-2023-multijugate,su-etal-2022-multi,hua-etal-2023-improving,li-etal-2023-dionysus,sun-etal-2022-multimodal,moghe-etal-2023-multi3nlu,bhat-etal-2023-adversarial,de-vries-etal-2022-make,eder-etal-2021-anchor,zhang-etal-2019-improving-low,fang-cohn-2017-model,xia-etal-2019-generalized,liu-etal-2023-investigating,schlichtkrull-sogaard-2017-cross,dingliwal-etal-2021-shot,ebrahimi-etal-2023-meeting,rottger-etal-2022-data,ghosh-etal-2023-dale,ding-etal-2020-daga,zou-etal-2021-low,lux-vu-2022-language,zheng-etal-2021-wav-bert,liu-etal-2023-vit}]
\subparagraph{Technology}
[{\cite{bamutura-etal-2020-towards,byamugisha-2022-noun,melgarejo-etal-2022-wordnet,yimam-etal-2020-exploring,himoro-pareja-lora-2022-preliminary,li-etal-2022-multi-level,niyongabo-etal-2020-kinnews,duggenpudi-etal-2022-teluguner,avram-etal-2022-distilling,Lane2021Nov,eskander-etal-2020-morphagram,rocha-souza-etal-2020-identification,lane-bird-2020-interactive,de-lhoneux-etal-2022-zero,imanigooghari-etal-2022-graph,brixey-etal-2020-exploring,rijhwani-etal-2020-soft,sikasote-anastasopoulos-2022-bembaspeech,adouane-etal-2020-identifying,botha-etal-2020-entity,moeller-etal-2021-pos,jin-etal-2020-unsupervised,dhar-etal-2022-evaluating,pfeiffer-etal-2020-mad,leong-etal-2022-bloom}]
\end{document}